\documentclass{article}

\usepackage[hide]{ed}
\usepackage{xspace}
\usepackage{url,wrapfig}
\usepackage{amsfonts}

\newtheorem{example}{Example}[section]

\newcommand{\hide}[1]{} 
\newcommand{\weg}[1]{}

\usepackage{graphicx}

\usepackage{lstsemantic}
\usepackage{paralist}

\newcommand{\OWL}{\ensuremath{\mathsf{OWL}}\xspace}

\newcommand{\RDF}{\ensuremath{\mathsf{RDF}}\xspace}
\newcommand*{\DOL}{\ensuremath{\mathsf{DOL}}\xspace}

\newcommand*{\CL}{\ensuremath{\mathsf{CL}}\xspace}
\newcommand{\CLminus}{\CL{}$^-$}

\newcommand{\SROIQD}{\ensuremath{\mathcal{SROIQ}(D)}\xspace}

 \newcommand{\Protege}{\textmd{\textsc{prot\'eg\'e}}\xspace }

\newcommand{\Sig}[1]{\ensuremath{\mathsf{Sig}}(#1)}

\usepackage[T1]{fontenc}
\usepackage{times}%

\title{Ontohub: A semantic repository for heterogeneous ontologies}

\author{Mihai Codescu$^1$, Eugen Kuksa$^2$, Oliver Kutz$^1$,\\
 Till Mossakowski$^2$, Fabian Neuhaus$^2$\\
 \small
$^1$ Free University of Bozen-Bolzano, Italy, \\
\small
E-mail: \{Mihai.Codescu,Oliver.Kutz\}@unibz.it \\
\small
$^2$ Otto-von-Guericke University of Magdeburg, Germany, \\
\small
E-mail: \{kuksa,till,fneuhaus\}@iks.cs.ovgu.de
}

\begin{document}
\maketitle

\begin{abstract}
  Ontohub is a repository engine for managing distributed
  heterogeneous ontologies. The distributed nature enables communities
  to share and exchange their contributions easily. The heterogeneous
  nature makes it possible to integrate ontologies written in various
  ontology languages.  Ontohub supports a wide range of formal logical and
  ontology languages, as well as various structuring and
  modularity constructs and inter-theory (concept) mappings, building
  on the OMG-standardized DOL language. Ontohub repositories are 
  organised as Git repositories, thus inheriting all features of
  this popular version control system.
  Moreover, Ontohub is the first repository engine meeting a
  substantial amount of the requirements formulated in the context of the Open
  Ontology Repository (OOR) initiative, including an API
  for federation as well as support for logical inference
  and axiom selection.
\end{abstract}



\section{Introduction}\label{sec:intro}

Ontologies play a central role for enriching data with a conceptual
semantics and hence form an important backbone of the Semantic
Web. The number of ontologies that are being built or already in
use is steadily growing.  This means that facilities for organizing
ontologies into repositories, searching, maintenance and so on are
becoming more important.

  Ontohub's overall design was conceived to satisfy a substantial subset of the requirements set out in the  Open Ontology Repository (OOR) initiative.\footnote{See \url{http://www.oor.net}} 
  OOR is a long-term international effort, which established requirements and designed an overall 
  architecture for ontology repositories. Ontohub is the first implementation of a repository engine meeting a substantial amount of OOR's requirement, including an API for federation as well as support for logical inference and axiom selection. Moreover, it extends the initial OOR vision by several features suggested by the development of the DOL language. First and foremost, this includes the fundamental abstraction from particular ontology languages, providing a principled logic-based support for heterogeneity in ontology design, based on general model-theoretic semantics. 
  
\ednote{@Oliver: extend a bit}

This paper is an updated and extended version of \cite{DACS2014}. Among the 
new case studies presented here, we include one that appeared in \cite{DBLP:conf/fois/KuksaM16}.
In the next two subsections, we will give an overview of Ontohub's features
and discuss related work. Section~\ref{sec:dol} introduces  
the Distributed Ontology Language (DOL) that is central to Ontohub. 
Ontohub's architecture is described in Sect.~\ref{sec:arch}.
Section~\ref{sec:case-studies} shows Ontohub at use with some ontology
alignments, as well as with theorem proving. Section~\ref{sec:concl}
concludes the paper.

\subsection{Features of Ontohub}\label{sec:ontohub}

Ontohub is a novel web-based repository engine. Central features are:
\begin{description}
  \item[multiple repositories] ontologies can be organized in multiple
    repositories, each with its own management of editing and
    ownership rights,
  \item[Git interface] version control of ontologies is supported via
    interfacing the Git version control system,
  \item[linked-data compliance] one and the same URL is used for
    referencing an ontology, downloading it (for use with tools), and
    for user-friendly presentation in the browser.
\end{description}

Ontohub is unique in
following OOR's ambitious goals, in particular, 
\begin{description}
  \item[modular architecture] Ontohub is decoupled into different services,
  \item[multi-language support] ontologies can formalized in various logics like OWL, Common Logic, TPTP and higher-order logic,
  \item[logical inference] intended consequences of ontologies can be proved.
\end{description}
Finally, Ontohub fully supports \textbf{modular and
  distributed ontologies} through the Distributed Ontology Language
(DOL), a standard approved by Object Management Group (OMG)
\cite{womo13,DOL-OMG,JYB-Festschrift2015-DOL}, see also
\url{dol-omg.org}.  DOL provides a unified framework
for \begin{inparaenum}[(1)]
  \item heterogeneous ontologies formalized in more than one logic,
  \item modular ontologies,
  \item mappings between ontologies (alignments, interpretation of theories, conservative
  extensions, translation to other ontology languages etc.).
\end{inparaenum}
All of these features are equipped with a formal semantics.

Users of Ontohub can upload, browse, search and annotate basic
ontologies in various languages via a web frontend, see
\url{https://ontohub.org}.  Ontohub is open source under GNU AGPL 3.0
license, the sources are available at the following URL
\url{https://github.com/ontohub/ontohub}.  Currently, Ontohub has about 200
registered users, which include ontology
researchers, ontology developers as well as master and PhD students.

\begin{figure}[h!]
  \centering
  \includegraphics[width=0.9\textwidth]{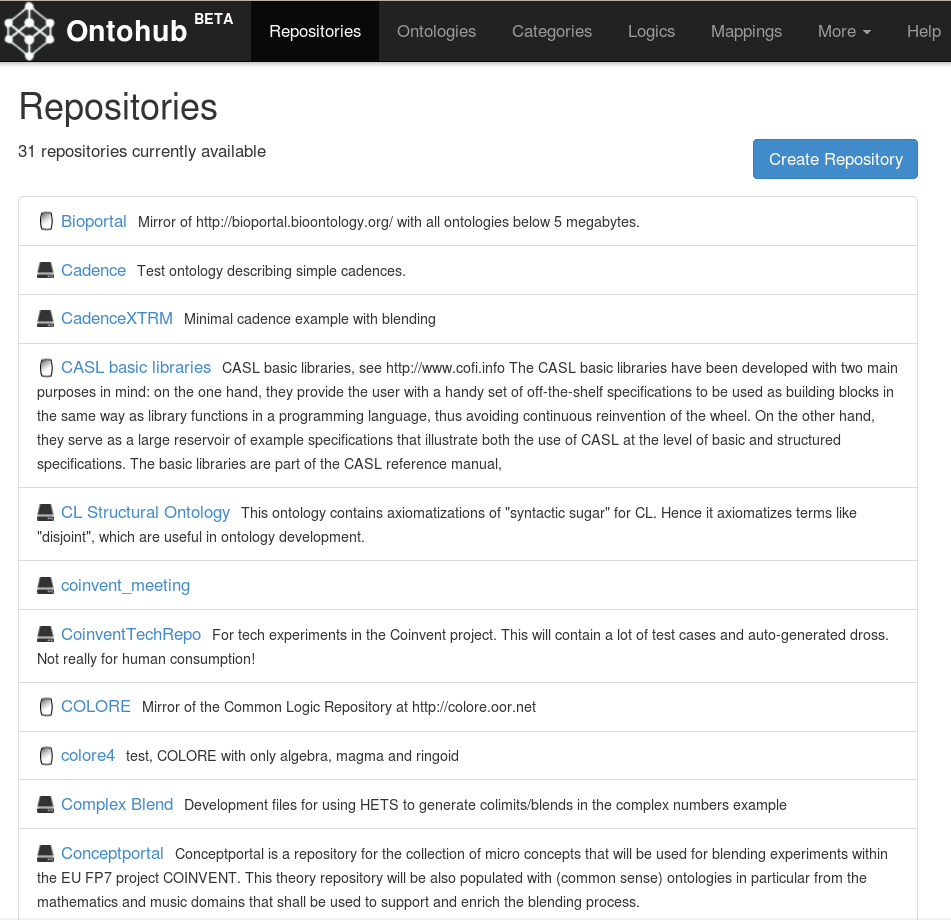}
  \caption{Overview of Ontohub repositories}
  \label{fig:ontohub-repos}
\end{figure}

Ontohub is not a repository, but a semantic repository engine. This means that
Ontohub ontologies are organized into repositories. See
Fig.~\ref{fig:ontohub-repos} for an overview of the currently
available repositories. Some of them, e.g.\ Bioportal or COLORE, are
mirrors of repositories hosted elsewhere (as indicated with the mirror
icons), while the others are native Ontohub repositories.  The
organisation into repositories has several advantages:
\begin{itemize}
\item
 Firstly, repositories provide a certain structuring of ontologies,
 let it be thematically or organisational. Access rights can be given
 to users or teams of users per repository. Typically, read access is
 given to everyone, and write access only to a restricted set of users
 and teams. However, also completely open world-writeable repositories
 are possible, as well as private repositories visible only to a
 restricted set of users and teams.  Since creation of repositories is
 done easily with a few clicks, this supports a policy of many but
 small repositories (which of course does not preclude the existence
 of very large repositories). Note that also structuring within
 repositories is possible, since each repository is a complete file
 system tree.
 
\item
 Secondly, repositories are Git repositories. Git is a popular
 decentralised version control system. With any Git client, the user
 can clone a repository to her local hard disk, edit it
 with any editor, and push the changes back to Ontohub. Alternatively,
 the web frontend can be used directly to edit ontologies; pushing
 will then be done automatically in the background. Parallel edits of
 the same file are synchronized and merged via Git; handling of
 merge conflicts can be done with Git merge tools.
\item
Thirdly, ontologies can be searched globally in Ontohub, or in
specific repositories. Additionally, user-supplied metadata 
can be used for searching.
\end{itemize}

Ontohub is linked-data compliant. This means that ontologies are
referenced by a unique URL of the form
\url{https://ontohub.org/name-of-repository/path-within-repository}.\footnote{In the future, this may change into
\url{https://ontohub.org/account-name/name-of-repository/path-within-repository}.} 
Depending on the MIME type of the request, under this URL, the raw
ontology file will be available, but also a HTML version for display
in a browser and a JSON version for processing with tools.

\subsection{Related Work}

Existing ontology resources on the web include search engines like
Swoogle, Watson, and Sindice. They concentrate on (full-text and
structured) search and querying. Ontology repositories also provide
persistent storage and maintenance.  TONES \cite{Tones} is a
repository for OWL \cite{SROIQ} ontologies that provides some metrics,
as well as an OWL sublanguage analysis. BioPortal \cite{BioPortal} is
a repository that originates in the biomedical domain, but now has
instances for various domains. Beyond browsing and searching, it
provides means for commenting and aligning ontologies. Besides OWL,
also related languages like OBO \cite{smith2007obo} are supported.
The NeOn Toolkit \cite{NeOn} supports searching, selecting, comparing,
transforming, aligning and integrating ontologies. It is based on the
OWL API and is no longer actively maintained.

\begin{figure}
  \centering
  \includegraphics[width=0.7\textwidth]{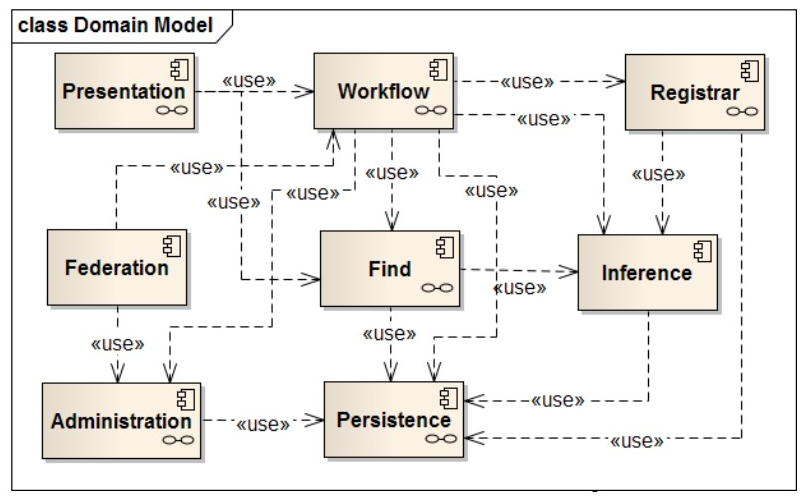}
  \caption{Architecture of the Open Ontology Repository (OOR)}
  \label{fig:arch-oor}
\end{figure}

The Open Ontology Repository (OOR) initiative aims at ``promot[ing]
the global use and sharing of ontologies
by \begin{inparaenum}[(i)]\item establishing a hosted
  registry-repository; \item enabling and facilitating open,
  federated, collaborative ontology repositories, and \item
  establishing best practices for expressing interoperable ontology
  and taxonomy work in registry-repositories\end{inparaenum}, where an
  ontology repository is a facility where ontologies and related
  information artifacts can be stored, retrieved and
  managed''~\cite{OOR:webpage}. One important goal of OOR is the
  support of ontology languages beyond OWL, for example Common Logic.
  Another goal is the support of logical inference.  Some
  proposed architecture is shown in Fig.~\ref{fig:arch-oor}.
  OOR is a
  long-term initiative, which has not resulted in a complete
  implementation so far\footnote{The main implementation used by OOR
    is (a cosmetically adapted) BioPortal, which however does not
    follow the OOR principles very much. There are no OOR
    implementation efforts beyond Ontohub.}, but established
  requirements and designed an architecture.\footnote{See
    \url{http://ontologforum.org/index.php/OpenOntologyRepository_Requirement}
and \url{http://ontolog.cim3.net/wiki/OpenOntologyRepository_Architecture/Candidate03.html}.}

\section{DOL}\label{sec:dol}

\begin{wrapfigure}{l}{0.6\textwidth}
               \centering
\vspace{-5mm}
        \scalebox{0.7}{
  \includegraphics[width=0.8\textwidth]{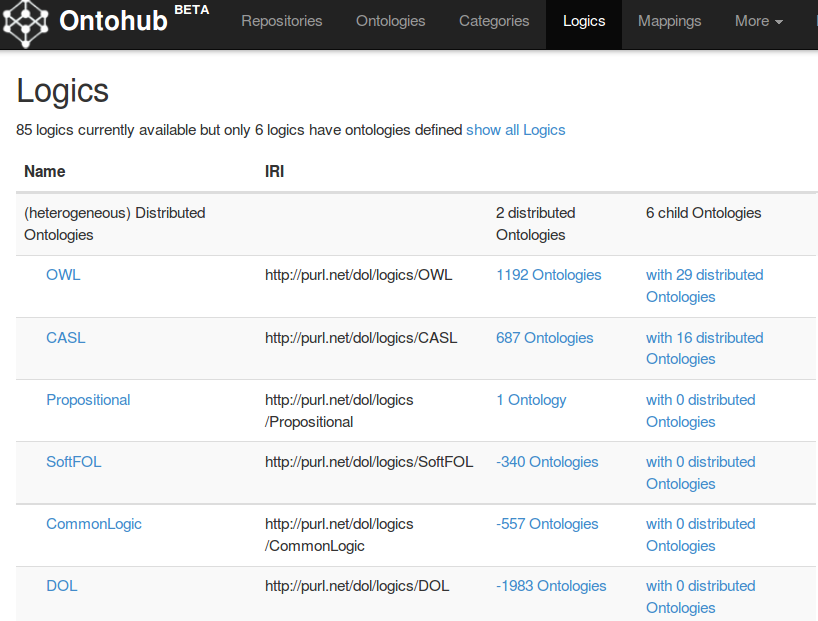}
}
  \caption{\texttt{ontohub.org} portal: overview of logics}
  \label{fig:ontohub-logics}
\end{wrapfigure}
The modularity mechanisms of Ontohub are based on those of
the Distributed Ontology Language (DOL).
DOL aims
at providing a unified framework for \begin{inparaenum}[(1)]
  \item ontologies formalized in heterogeneous logics,
  \item modular ontologies,
  \item links between ontologies, and
  \item annotation of ontologies.
\end{inparaenum}


\subsection{Logics}\label{sec:logics}

The large variety of logics in use can be captured at an abstract level using the concept of logic syntax, which we introduce below.
This allows us to develop results independently of the particularities of a logical system. The main idea is to collect the non-logical
symbols of the language in signatures and to assign to each signature the set of sentences that can be formed with its symbols. 
For each signature, we provide means for extracting the symbols it consists of, together with their kind.
Signature morphisms are mappings between signatures. 
Formally, signatures and their morphisms form a so-called category, which can be
understood as a graph together with a composition principle that identifies
paths in the graph.
Signature morphisms induce a mapping 
between the sentences of the signatures, usually by replacing the symbols that occur
in a sentence with their image through the signature morphism.
\cite{JYB-Festschrift2015-DOL} provides full details of the formal aspects of DOL.


A logic syntax can be complemented with a model theory, which introduces semantics for the language and 
gives a satisfaction relation between the models and the sentences of a signature. The result is a so-called \emph{institution}
 \cite{Ins}. Similarly, we can complement a logic syntax with a proof theory, introducing a derivability relation
between sentences, thus obtaining an \emph{entailment system} \cite{Meseguer:gl89}. In particular, this can be 
done for all logics in use in Ontohub.

\begin{example}
\OWL signatures consist of sets of atomic classes, individuals and properties. 
\OWL signature morphisms map classes to classes, individuals 
to individuals, and properties to properties. For an \OWL signature $\Sigma$, sentences are subsumption relations between classes, membership
assertions of individuals on classes and pairs of individuals in properties. Sentence translation along a signature morphism is simply
replacement of non-logical symbols with their image along the morphism. 
The kinds of symbols are class, individual, object property and data property, respectively, and the set of symbols of a signature is the union of its sets of
classes, individuals and properties.
\end{example}

In this framework, an ontology $O$ over a logic syntax $L$ is a pair $(\Sigma, E)$ where $\Sigma$ is a signature and $E$ is a set of
$\Sigma$-sentences. Given an ontology $O$, we denote by $\Sig{O}$ the signature of the ontology. An ontology morphism
$\sigma : (\Sigma_1, E_1) \rightarrow (\Sigma_2, E_2)$ is a signature morphism $\sigma: \Sigma_1 \rightarrow \Sigma_2$ such that
$\sigma(E_1)$ is a logical consequence of $E_2$.
\begin{figure}[htp]
  \includegraphics[width=0.95\textwidth]{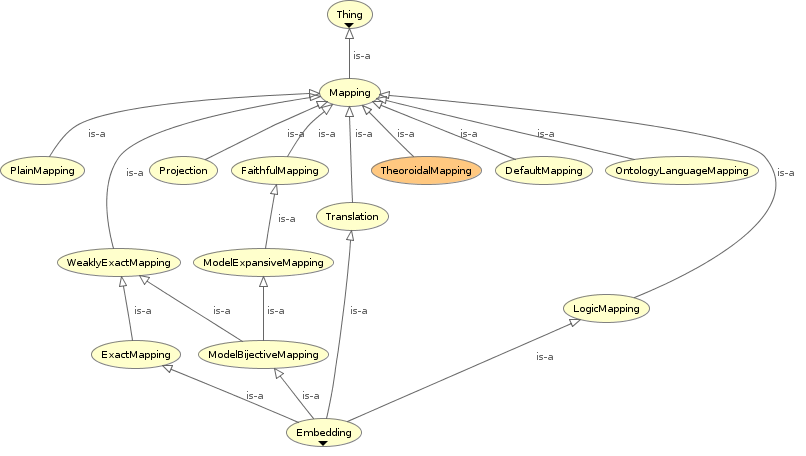}  
  \caption{The part of the DOL ontology concerning mappings}
  \label{fig:mappings}
\end{figure}
Several notions of \emph{translations} between logics can be
introduced. In the case of logic syntaxes, the simplest variant of
translation from $L_1$ to $L_2$ maps $L_1$-signatures along $\Phi$ to
$L_2$-signatures and $\Sigma$-sentences in
$L_1$ to $\Phi(\Sigma)$-sentences in $L_2$, for each $L_1$-signature
$\Sigma$, in a compatible way with the sentence translations along
morphisms. The complexity of translation increases when a model theory
or a proof theory is added to the logic syntax.
Fig.~\ref{fig:mappings} shows the inferred class hierarchy below the
class \texttt{Mapping} of the DOL ontology (see
Sect.~\ref{sec:registry} below), as computed within \Protege. 
Mappings are split along the following dichotomies:
\begin{itemize}
\item \emph{translation} versus \emph{projection}: a translation embeds or encodes a logic into another one, while a projection is a forgetful operation (e.g.\ the projection from first-order logic to propositional logic forgets predicates with arity greater than zero). Technically, the distinction is that between institution comorphisms and morphisms \cite{GoguenRosu02}.
\item \emph{plain mapping} versus \emph{simple theoroidal mapping} \cite{GoguenRosu02}: while
a plain mapping needs to map signatures to signatures, a simple theoroidal mapping maps signatures to theories. The latter therefore allows for using
``infrastructure axioms'': e.g.\ when mapping \OWL to Common Logic,
it is convenient to rely on a first-order axiomatization of
a transitivity predicate for properties etc.\
\end{itemize}

Mappings can also be classified according to their accuracy, see
\cite{MoKu:OntoGraph11} for details. 
\emph{Sublogics} are the most accurate
mappings: they are just syntactic subsets. \emph{Embeddings} come close to
sublogics, like injective functions come close to subsets. A mapping can be \emph{faithful} in the sense that logical
consequence (or logical deduction) is preserved and reflected, that is, inference systems and
engines for the target logic can be reused for the source logic
(along the mapping). \emph{(Weak) exactness} is a technical property that
guarantees this faithfulness even in the presences of ontology
structuring operations \cite{Borzyszkowski02}.

\subsection{A Graph of Logic Translations}

\begin{figure}[htp]
  \includegraphics[width=\textwidth]{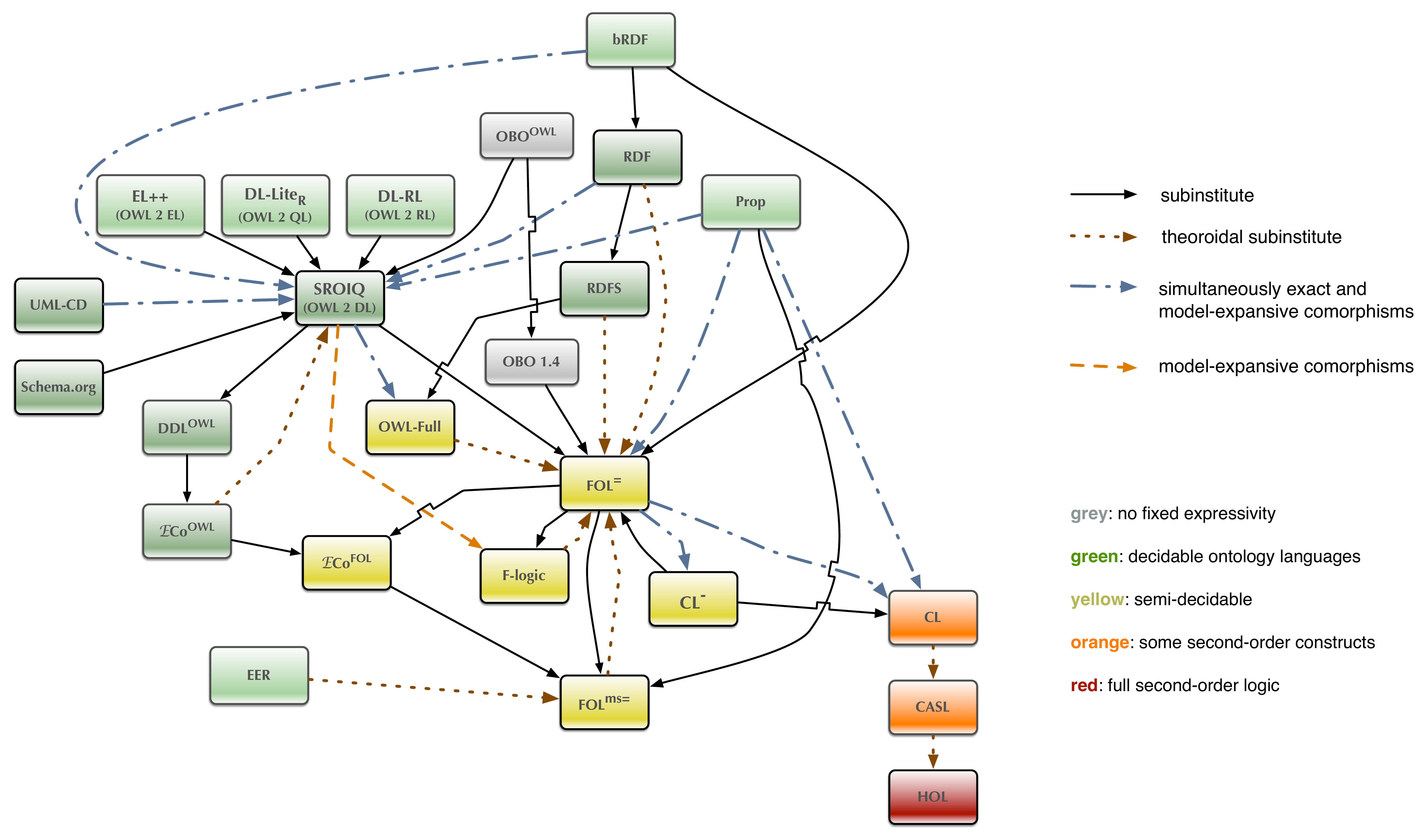}
  \caption{The logic translation graph for DOL-conforming languages}
  \label{fig:coregraph}
\end{figure}
Fig.~\ref{fig:coregraph} is a revised and extended version of the graph of logics and translations introduced in~\cite{MoKu:OntoGraph11}.  New nodes include UML class diagrams, \OWL-Full (i.e.\ \OWL with an RDF semantics instead of description logic semantics), and Common Logic without second-order features (\CLminus).  We have defined the translations between all of these logics in earlier publications~\cite{DOL3semantics,MoKu:OntoGraph11
}.  The definitions of the DOL-conformance of some central standard ontology languages and translations among them has been given as annexes to the standard, whereas the majority will be maintained in an open registry (cf.\ Sec.~\ref{sec:registry}).

\subsection{A Registry for Ontology Languages and Mappings}
\label{sec:registry}

The DOL standard is not limited to a fixed set of ontology
languages.  It will be possible to use any (future) logic or mapping
(in the sense of Sect.~\ref{sec:logics}) with DOL.  This led to
the idea of setting up a \emph{registry} to which the community can
contribute descriptions of any logics and mappings. Moreover, logics
can support ontology languages (e.g.\ \SROIQD \cite{SROIQ} supports
\OWL), which can in turn have different serializations.  All these
notions are part of the \DOL ontology. This ontology turns Ontohub itself
into part of the Semantic Web: it is mostly written in \RDF (the data
part) and \OWL (the concepts), but also contains first-order parts. We
use \RDF and \OWL reasoners in order to derive new facts in the DOL
ontology. A
full description and discussion of the DOL ontology can be found in
\cite{LangeEtAL12a}.

\begin{figure}[htp]
  \includegraphics[width=\textwidth]{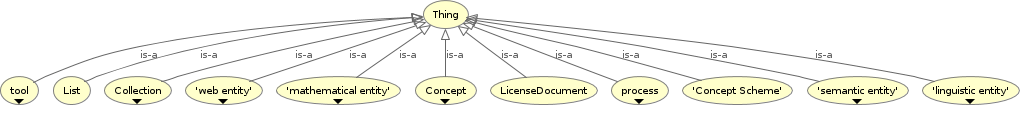}  
  \caption{Top-level classes in the \DOL ontology}
  \label{fig:top-level-classes}
\end{figure}
Fig.~\ref{fig:top-level-classes} shows the
top-level classes of the DOL ontology's \OWL module, axiomatising logics, languages, and mappings to the extent possible in \OWL.
Object-level classes (that is, classes providing the vocabulary
for expressing distributed ontologies) comprise ontologies, their constituents (namely
entities, such as classes and object properties, and sentences, such as
class subsumptions), as well as links between
ontologies.

Mappings are modelled by a hierarchy of properties corresponding to
the different types of edges in Fig.~\ref{fig:coregraph}; see also 
Fig.~\ref{fig:mappings}.   
The full DOL ontology is available at \url{http://www.omg.org/spec/DOL/DOL-terms.rdf}.

\section{Architecture of Ontohub}\label{sec:arch}

\begin{figure}
  \centering
  \includegraphics[width=0.75\textwidth]{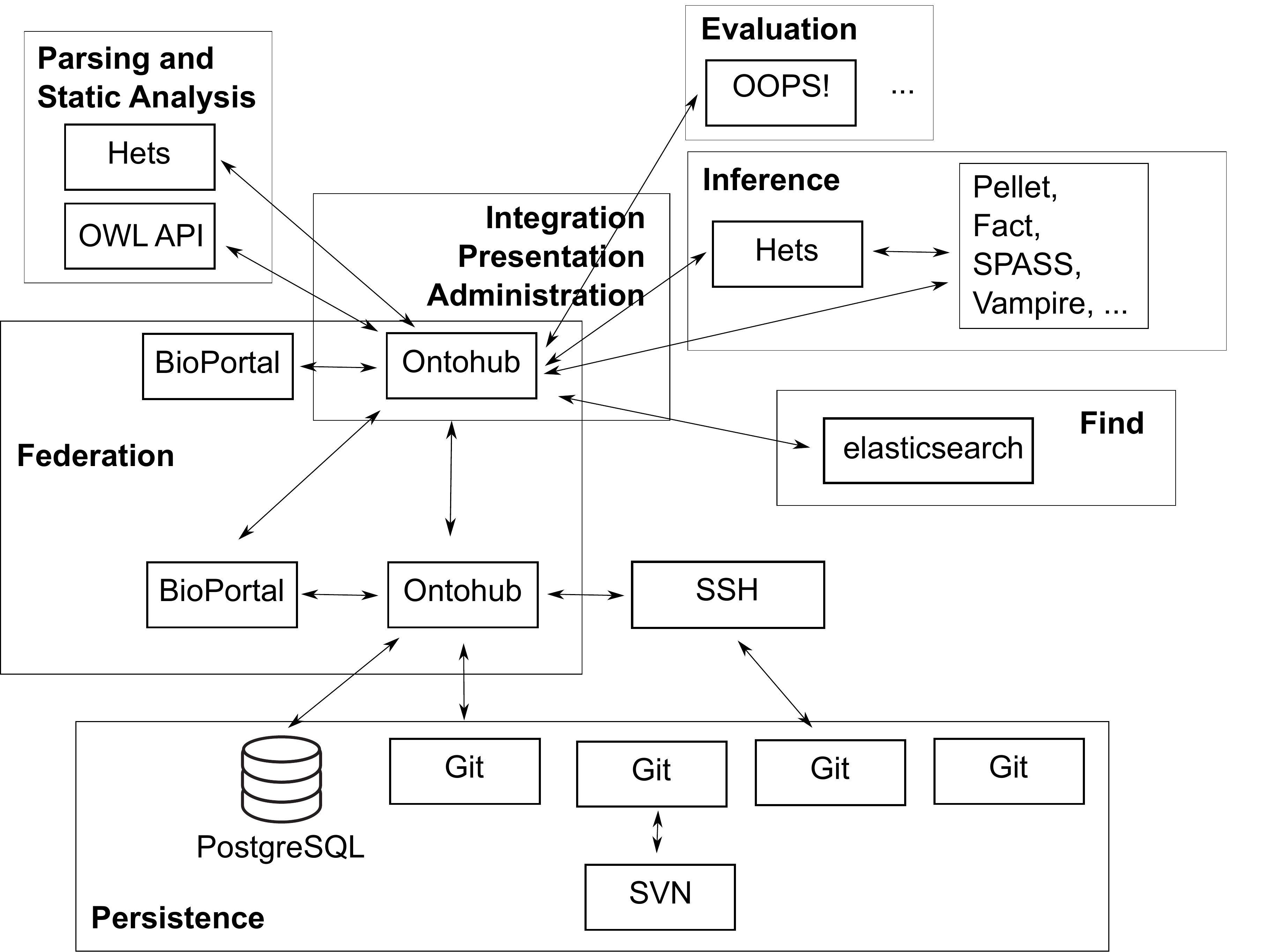}
  \caption{Ontohub in a network of web services}
  \label{fig:arch-ontohub}
\end{figure}

Fig.~\ref{fig:arch-ontohub} depicts the  Ontohub architecture.
The most challenging part of Ontohub's implementation is the complex
tool integration.
The key feature of the OOR architecture is the decoupling into
decentralised services, which are ontologically described (thus
arriving at Semantic Web services).  With Ontohub, we are moving towards this
architecture, while keeping a running and usable system.
We now briefly describe these services.

The services are centrally integrated by the Ontohub \emph{integration}
layer, which is a  Ruby on Rails application that also includes the
\emph{presentation} layer, i.e.\ a front-end providing the web
interface, as well as the \emph{administration} layer,
i.e.\ user rights management and authorisation.


The \emph{persistence} layer is based on Git (via git-svn, also
Subversion repositories can be used) and an SQL database.  The
database backend is PostgreSQL. For the Git
integration into the web application, a custom Git client was
implemented in Ruby to be less prone to errors due to changes in new
versions of the official Git command line client.

Efficient indexing and searching (the \emph{find} layer) is done via  
elasticsearch.

A \emph{federation} API allows the data exchange among Ontohub
and also with BioPortal instances. We therefore have generalised the
OWL-based BioPortal API to arbitrary ontology languages, e.g.\ by
abstracting classes and object properties to symbols of various kinds.

\emph{Parsing and static analysis} is a RESTful service of its own provided by
the Heterogeneous Tool Set (Hets~\cite{MossakowskiEA06},
available at \url{http://hets.eu}). Hets supports a large number of
basic ontology languages and logics and is capable of describing the
structural outline of an ontology from the perspective of DOL, which
is not committed to one particular logic.  Hets returns the symbols
and sentences of an ontology in XML format. Hets can do this for a
large variety of ontology languages, while the OWL API does scale
better for very large OWL ontologies.  The latter is an example for a
service of Ontohub which is provided for a restricted set of ontology
languages.

We have integrated OOPS! \cite{poveda2012validating}
as an ontology \emph{evaluation} service (for OWL only), and from the
OOPS! API, we have derived a generalised API for use with other
evaluation services.

\emph{Inference} is done by encapsulating standard
batch-processing reasoners (Pellet, Fact, SPASS, Vampire etc.) into a
RESTful API through Hets (which has been interfaced with
15 different reasoners).
Integrating support for logical inference required a substantial
extension of Hets's HTTP interface which returns proof details in JSON format.
The prover-independent implementation of the SInE algorithm is a novelty in this field since it has only been used with few provers so far.
In Ontohub, it operates independently of the prover and, thus, supports any prover available in Ontohub.

\section{Case Studies} \label{sec:case-studies}

\subsection{Ontology alignment in Ontohub}

The foundational ontology (FO) repository Repository of Ontologies for MULtiple USes (ROMULUS)\footnote{See \url{http://www.thezfiles.co.za/ROMULUS/home.html}}
contains alignments between a number of foundational ontologies, expressing semantic relations between the aligned
entities. We select three such ontologies, containing spatial and temporal concepts: DOLCE\footnote{See \url{http://www.loa.istc.cnr.it/DOLCE.html}}, GFO\footnote{See \url{http://www.onto-med.de/ontologies/gfo/}} and BFO\footnote{See \url{http://www.ifomis.org/bfo/}}, and present alignments between them 
using \DOL syntax:\footnote{This and the other examples from this section can be found at:\\ \url{https://ontohub.org/repositories/ontohubaopaperexamples}}


\begin{lstlisting}[basicstyle=\ttfamily\footnotesize,language=dolText,morekeywords={props,ObjectProperty,Class,DisjointUnionOf,SubClassOf,Characteristics,Transitive,Asymmetric,SubPropertyOf,DisjointClasses,EquivalentTo,inverse,only,forall,iff,if,or,exists,bridge,distributed,from},escapechar=@,mathescape]
%prefix(
           gfo: <http://www.onto-med.de/ontologies/>
           dolce: <http://www.loa-cnr.it/ontologies/>
           bfo: <http://www.ifomis.org/bfo/>
           
        )%
logic OWL

alignment DolceLite2BFO :
  dolce:DOLCE-Lite.owl
  to
  bfo:1.1 =
 endurant = IndependentContinuant,
 physical-endurant = MaterialEntity,
 physical-object = Object,   perdurant = Occurrent,
 process = Process,          quality = Quality,
 spatio-temporal-region = SpatiotemporalRegion,
 temporal-region = TemporalRegion,  space-region = SpatialRegion

alignment DolceLite2GFO :
  dolce:DOLCE-Lite.owl to gfo:gfo.owl =
 	particular = Individual, endurant = Presential,
 	physical-object = Material_object, amount-of-matter = Amount_of_substrate,
 	perdurant = Occurrent, 	quality = Property,
 	time-interval = Chronoid, generic-dependent < necessary_for,
 	part < abstract_has_part, part-of < abstract_part_of,
 	proper-part  <	 has_proper_part,  	proper-part-of  < proper_part_of,
 	generic-location < occupies, 	generic-location-of < occupied_by

alignment BFO2GFO :
  bfo:1.1 to gfo:gfo.owl =
	Entity = Entity, Object = Material_object,
	ObjectBoundary  = Material_boundary, Role < Role ,
 	Occurrent = Occurrent, 	Process = Process, Quality = Property
 	SpatialRegion 	= Spatial_region, TemporalRegion = Temporal_region 	
\end{lstlisting}

We can then combine the ontologies while taking into account the semantic dependencies given by the alignments using
\DOL combinations:

\begin{lstlisting}[basicstyle=\ttfamily\footnotesize,language=dolText,morekeywords={props,ObjectProperty,Class,DisjointUnionOf,SubClassOf,Characteristics,Transitive,Asymmetric,SubPropertyOf,DisjointClasses,EquivalentTo,inverse,only,forall,iff,if,or,exists,bridge,distributed,from},escapechar=@,mathescape]
ontology Space =
 combine BFO2GFO, DolceLite2GFO, DolceLite2BFO
\end{lstlisting}

\begin{figure}[h!]
  \centering
  \includegraphics[width=\textwidth]{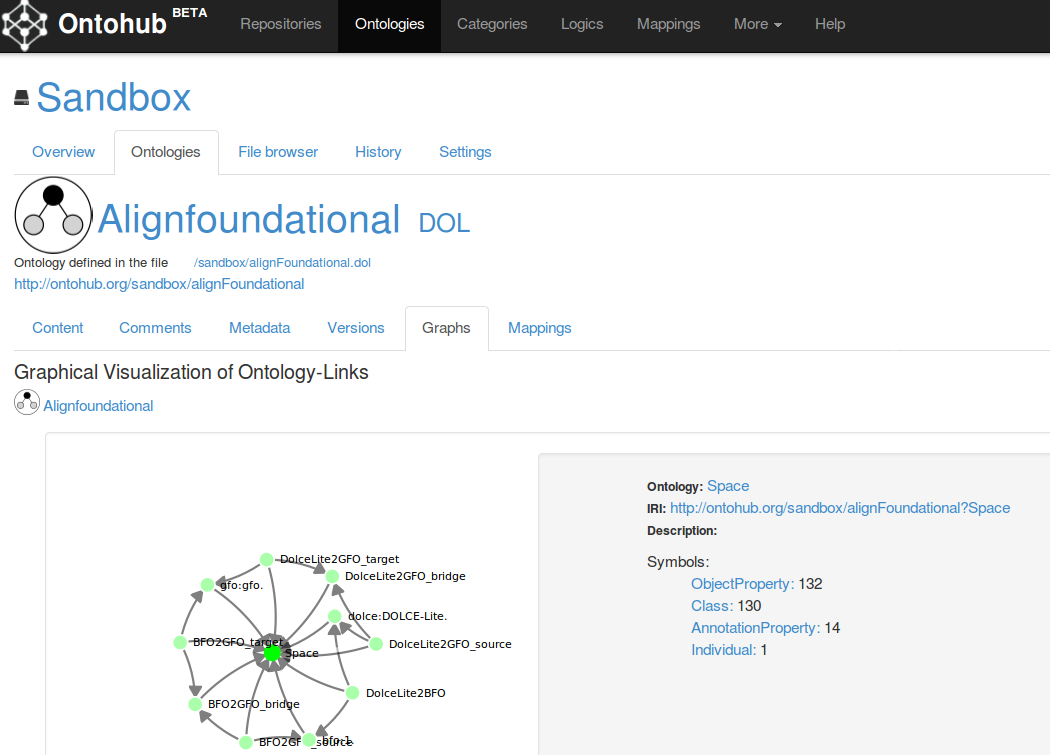}  
  \caption{Combination of ontologies along alignments.}
  \label{fig:colimit}
\end{figure}

Fig.~\ref{fig:colimit} shows the graph of links between ontologies created by Ontohub as a result of analysis of the 
{\tt Space} ontology, which appears in the center of the graph. Around it and linking to it there are the aligned ontologies
together with the diagrams resulting from the analysis of the alignments.\footnote{Details on the construction of these 
diagrams can be found in \cite{om2014}.}

\subsection{Ontology Competency Questions with Ontohub and DOL}
`Competency questions' is the name for a methodology, which supports behavior-driven development of ontologies. The approach can be, somewhat simplified,  summarized in the following steps \cite{gruninger1995methodology}:
\begin{enumerate}
	\item The use cases for the soon-to-be-developed ontology are captured in form of scenarios. Each scenario describes a possible state 
of the world and raises a set of competency questions. The answers to these competency questions should follow 
logically from the scenario -- 
provided the knowledge that is supposed to be represented in the ontology.  
	\item A scenario and its competency questions are formalized or an existing formalization is refined. 
	\item The ontology is developed. 
	\item An automatic theorem prover is used to check whether the competency questions logically follow from the scenario and the ontology. 
	\item Steps (2-4) are repeated until all competency questions can be proven from the combination of the ontology and their respective scenarios.
\end{enumerate}

Ontohub enables the representation and execution of competency questions with the help of 
DOL files. For example, let's assume we are planning to develop an ontology of family relationships called \emph{familyRel.omn}. One way to capture the intended capabilities of the ontology is the following:
\begin{center}
\begin{minipage}{0.9\textwidth}
\begin{itshape}
The use case is to enable semantically enhanced searches for a database, which contains names of people, their gender, and information about parenthood.  Assuming the database contains the following information: 
\begin{itemize}
	\item Amy is female and a parent of Berta and Chris.  
	\item Berta is female. 
	\item Chris is male and a parent of Dora. 
	\item Dora is female. 
\end{itemize}
In this case the system should be able to answer the following questions
	\begin{itemize}
		\item Is Chris a father of Dora? (expected: yes)
		\item Is Berta a sister of Chris (expected: yes)
		\item Is Chris female? (expected: no)
		\item Is Amy older than Dora? (expected: yes)
	\end{itemize}
\end{itshape}
\end{minipage}
 \end{center}

These competency questions can be encoded in DOL (see Fig. \ref{fig:cq-dol}). Ontohub analyses the DOL file and recognises the proof obligations
that are derived from the competency questions. These proof obligations can be validated with  theorem proving (see section \ref{sec:thproving}). 

 \begin{figure}[h]
	   \centering
\begin{lstlisting}[basicstyle=\ttfamily\footnotesize,language=dolText,morekeywords={props,ObjectProperty,Class,DisjointUnionOf,SubClassOf,Characteristics,Transitive,Asymmetric,SubPropertyOf,DisjointClasses,EquivalentTo,inverse,only,forall,iff,if,or,exists,bridge,distributed,from},escapechar=@,mathescape]
%prefix( f1:   <http://ex.com/fr1#> )%

logic OWL

ontology scenario = <https://ontohub.org/appliedontologyontohubpaper/scenario> 
ontology genealogy = <https://ontohub.org/appliedontologyontohubpaper/familyRelations>

ontology CQbase =  genealogy  and scenario end

%%  Is Chris a father? (expected: yes)
ontology chrisFather =  CQbase then %implies 
  {  Individual: f1:Chris  Types: f1:Father } end

%%  Is Dora a child of Chris (expected: yes)
ontology doraChildChris =  CQbase then %implies 
  {  Individual: f1:Dora   Facts: f1:child_of f1:Chris } end

%%  Is Chris female? (expected: no)	
ontology chrisFemale =  CQbase then %implies 
  {  Individual: f1:Chris  Types: not f1:Female } end

%%  Is Amy older than Dora? (expected: yes)
ontology amyOlderDora =  CQbase then %implies 
  {  Individual: f1:Amy   Facts: f1:older_than f1:Dora } end
\end{lstlisting}
	     \label{fig:cq-dol}
\end{figure}

%

%
%

\begin{figure}[b]
  \centering
  \includegraphics[width=0.6\textwidth]{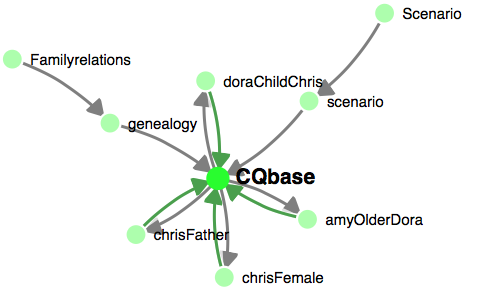}
  \caption{Representation of the Competency Questions in Ontohub}
  \label{fig:cq-graph}
\end{figure}

\subsection{Theorem Proving in Ontohub}\label{sec:thproving}

Ontohub recognises proof obligations in ontologies, like the competency questions above, and allows the user to invoke automated theorem provers to attempt to prove these conjectures.
For simplicity, these conjectures are called ``theorems'' in the web application.

\begin{figure}[h!]
  \centering
  \includegraphics[width=0.9\textwidth]{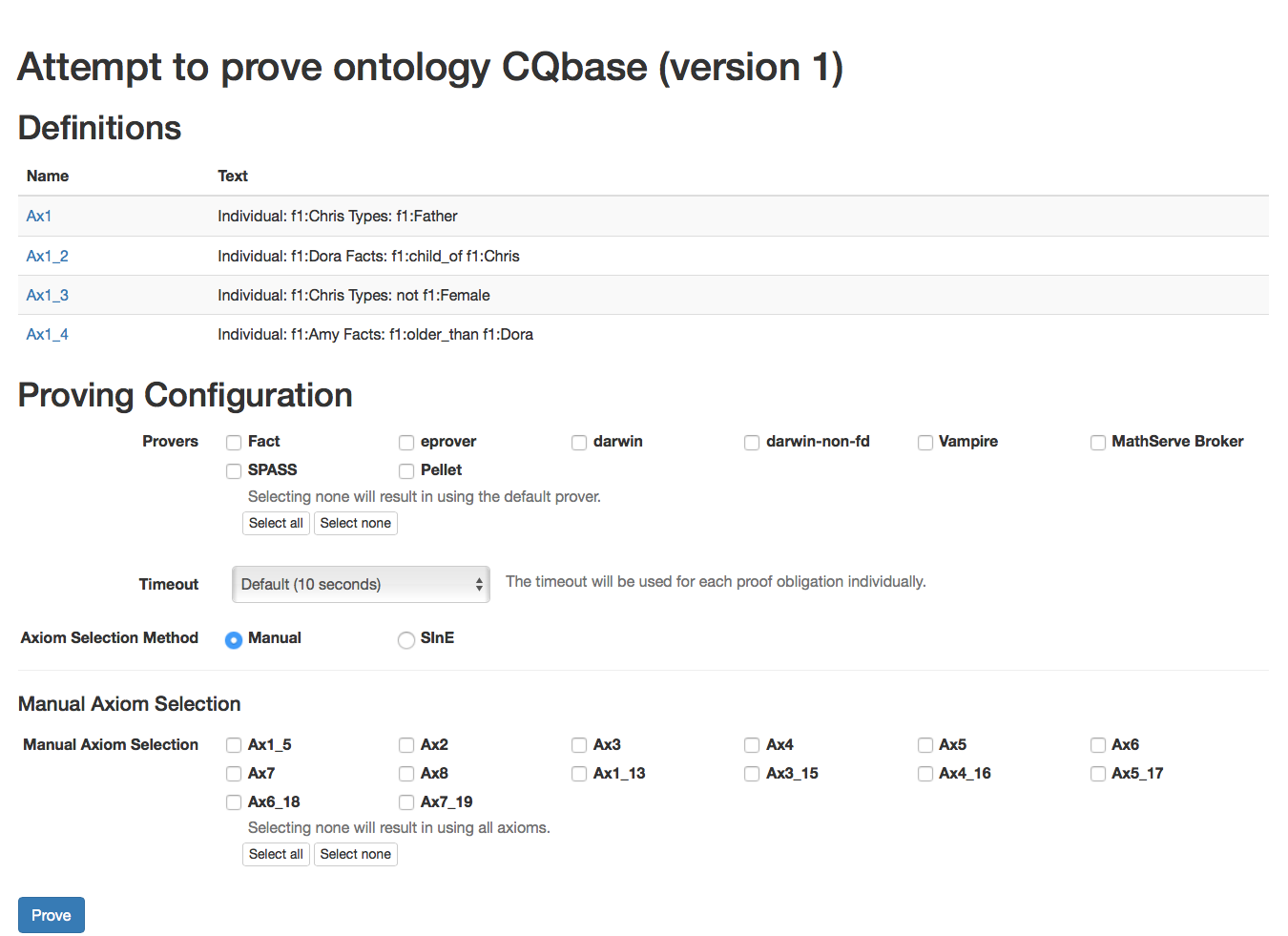}
  \caption{Configuration page for a proof attempt.}
  \label{fig:proof-attempt-configuration}
\end{figure}

When an ontology has been analysed, ``Theorems'' are shown in its ``Contents'' area.
There, the user can either choose to prove all conjectures at once or only a specific one.
Either way, the next step is to configure the proof attempts, as shown in Fig. \ref{fig:proof-attempt-configuration}.

Above of the actual configuration options, the selected ``theorems'' to be proved are listed with their names and their definitional text.
This can be, for example, the competency questions of the previous section.

Multiple \emph{provers} can be selected which are invoked in parallel for each selected conjecture.
If no prover is actively selected by the user, a default prover is used.

A \emph{timeout} for the automated theorem prover is used to limit the prover's resources.
When the given amount of time is exceeded and no proof or refutation has been found yet, the prover is stopped.
Another configuration may lead to a successful proof attempt.

\emph{Axiom selection} can be used to restrict a proof attempt to only include a subset of the ontology's axioms for proving.
This can reduce proving time and in some cases make proving feasible at all, which is particularly important for large ontologies.
Axioms can be selected manually or automatically with a heuristic.
The manual method allows the user to select every axiom that may be needed for a proof individually.
The automatic heuristic is a prover-independent implementation of the SInE algorithm~\cite{SInE}.
It expects three parameters to be set by the user that influence how many axioms are selected.
The configured axiom selection is shared between all the proof attempts resulting from this configuration.

\begin{figure}[h]
  \centering
  \includegraphics[width=0.9\textwidth]{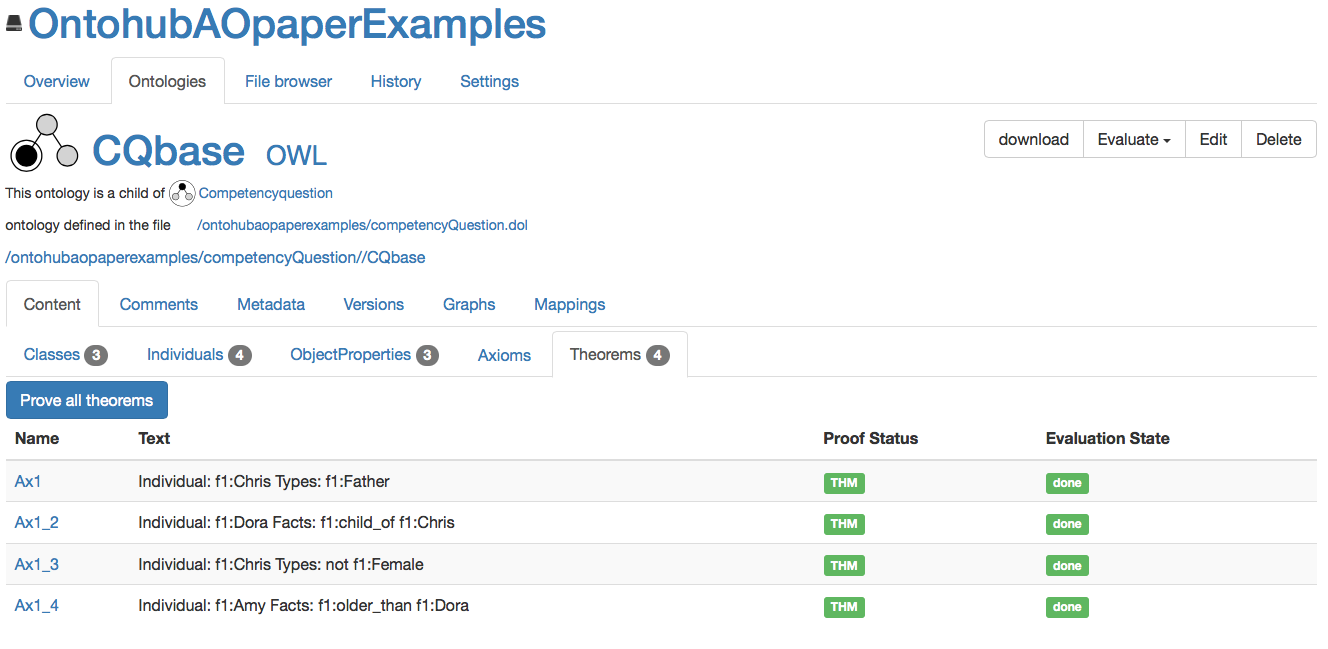}
  \caption{Overview of the statuses of all theorems. }
  \label{fig:proving_conjectures}
\end{figure}

When a proof attempt is finished, it is assigned a proof status telling the result in a single word, as displayed in Fig. \ref{fig:proving_conjectures}:
There, the first conjecture (stating that Chris is a Father) has been evaluated with the resulting status ``THM''. Thus, the ontology passes the corresponding competency question. 
These statuses are defined in the SZS ontology~\cite{SZS}.
The most common statuses used by provers are
\begin{enumerate}[(i)]
  \item THM (Theorem): All models of the axioms are models of the conjecture.
  \item CSA (CounterSatisfiable): Some models of the axioms are models of the conjecture's negation.
  \item TMO (Timeout): The prover exceeded the time limit.
\end{enumerate}
Of these statuses, ``THM'' and ``CSA'' indicate a successful prover run, while ``TMO'' shows that the prover did not finish successfully by exceeding the given amount of time.
We extended the SZS ontology ontology\footnote{Our modified SZS ontology can be found on \url{http://ontohub.org/meta/proof_statuses}.} by a status specifically for proving with reduced axiom sets:
\begin{enumerate}[(i)]
	\setcounter{enumi}{3}
  \item CSAS (CounterSatisfiableWithSubset): Some models of the selected axioms are models of the conjecture's negation.
\end{enumerate}
If a refutation of the conjecture is found using a strict subset of the axioms (which means that the prover returns with ``CSA''), we do not know whether the conjecture is really false or we have excluded an axiom that is crucial to a potentially existing proof.
If the prover returns with ``THM'', we know by monotonicity of entailment relations that the found proof is also a proof with the full axiom set.

Details of each proof attempt, including the proof itself if the attempt finished, can be inspected on the proof attempts page.
There, the user can see, for example, the configuration with the selected axioms and the actually used axioms.

\section{Conclusions and Future Work} \label{sec:concl}

Ontohub is on its way from a research prototype to
productive use.
The
FOIS 2014 ontology competition has used Ontohub as platform for
uploading ontologies used in submissions, see
\url{https://ontohub.org/fois-ontology-competition}. 
Ontologies used in FOIS papers often need expressiveness
beyond OWL; here, the multi-logic nature of Ontohub is essential. 
A good example for  these novel capabilities is given by the recent FOUST initiative ('FOundational STance'), an effort  to build a digital archive hosted on Ontohub to provide authoritative formalized versions of  the leading foundational ontologies (including BFO, DOLCE, GFO, GUM, UFO and YAMATO). This will include variants of these ontologies given e.g.\ in OWL and variants of FOL and HOL, as well as formally establishing their relationships based on theory interpretation or alignment. Another aim of the project is to provide consistency proofs, within Ontohub, by giving interpretations of the ontologies into formalised models.

Future work will improve stability and useability, and include the completion of full DOL support
and the integration of ontology evaluation and workflow tools.
The integration of interactive provers bears
many challenges; a first step is the integration of Isabelle via the
web interface Clide \cite{clide} developed by colleagues in Bremen,
which is currently equipped with an API for this
purpose. 

Currently, a re-implementation of Ontohub is under way. The idea is to
push forward the splitting into different services. Most prominently,
we plan to implement the Ontohub frontend completely in
ember.js/Javascript, while Ruby on Rails is only used for the JSON API served by the backend.
Communication with Hets, which is currently done by the background
process manager Sidekiq, will be done in the future through RabbitMQ,
which is a fully-fledged service broker. This will ease the
distribution of Hets parsing and proving services across server
farms.

\bibliographystyle{plain}  
\bibliography{ontohub,kwarc} 

\end{document}